\documentclass[a4paper,conference]{IEEEtran}
\usepackage{url}

\usepackage{caption}
\usepackage{subfig}
\usepackage{times}
\usepackage{epsfig}
\usepackage{graphicx}
\usepackage{amsmath}
\usepackage{amssymb}
\usepackage{float}

\ifCLASSINFOpdf
\else
\fi
\begin{document}
%
\title{Sliding Line Point Regression for Shape Robust Scene Text Detection}

\author{Yixing Zhu and Jun Du\\
	National Engineering Laboratory for Speech and Language Information Processing\\ University of Science and Technology of China\\
	Hefei, Anhui, China\\
	{\tt\small zyxsa@mail.ustc.edu.cn, jundu@ustc.edu.cn}
}


%


\maketitle

\begin{abstract}
Traditional text detection methods mostly focus on quadrangle text. In this study we propose a novel method named sliding line point regression (SLPR) in order to detect arbitrary-shape text in natural scene. SLPR regresses multiple points on the edge of text line and then utilizes these points to sketch the outlines of the text. The proposed SLPR can be adapted to many object detection architectures such as Faster R-CNN and R-FCN. Specifically, we first generate the smallest rectangular box including the text with region proposal network (RPN), then isometrically regress the points on the edge of text by using the vertically and horizontally sliding lines. To make full use of information and reduce redundancy, we calculate x-coordinate or y-coordinate of target point by the rectangular box position, and just regress the remaining y-coordinate or x-coordinate. Accordingly we can not only reduce the parameters of system, but also restrain the points which will generate more regular polygon. Our approach achieved competitive results on traditional ICDAR2015 Incidental Scene Text benchmark and curve text detection dataset CTW1500.
\end{abstract}


%
\IEEEpeerreviewmaketitle

\section{Introduction}

Text detection is important in our daily life as it can be applied in many areas, such as digitization of text, text translation, etc. In this study, we focus on scene text detection. Some of the previous methods \cite{zhong2016deeptext} \cite{liao2017textboxes} have obtained good results on many horizontal scene texts dataset based on Faster R-CNN \cite{gupta2016synthetic} or SSD \cite{liu2016ssd}. Some methods \cite{yin2015multi} \cite{kang2014orientation} \cite{liu2017deep} \cite{he2017deep} \cite{dai2017fused} \cite{zhou2017east} \cite{jiang2017r2cnn} also tried to solve arbitrary-oriented text detection problem. \cite{dai2017fused} and \cite{jiang2017r2cnn} regressed first a horizontal rectangle and then a quadrilateral. \cite{he2017sren} aimed to generate an irregular polygon after regressing a rectangle. The methods mentioned above mostly treated a text line as a quadrilateral which can be completely represented by four points. However, besides the quadrilateral shape, there are many other various shapes of text line in natural scene. Therefore, recent research \cite{yuliang2017detecting} \cite{kheng2017total} have begun to explore curve text line detection. In this paper we explore both arbitrary-oriented and curve text detection. Our method named sliding line point regression (SLPR) is based on 2-step object detection methods using Faster R-CNN or R-FCN. Firstly we propose some interesting rectangular regions with region proposal network (RPN), then regress the points on the edge of text. We generate some rules to determine which points should be regressed so that there will be relevance between points. Different from \cite{yuliang2017detecting} which directly regressed both x-coordinate and y-coordinate of fixed annotated points and employed RNN~\cite{graves2013speech,zhang2017gru} to learn their relevance, we introduce some rules to vertically and horizontally slide lines along text and then regress the intersection points of sliding lines and text lines as illustrated in Fig.~\ref{overview_img}. In this way, we can only regress x-coordinate or y-coordinate of these points, then calculate other coordinates with the position of rectangle, yielding reduction of unnecessary computation and improvement of performance.

The contributions of this paper are as follows:
\begin{figure}
	\begin{center}
       \includegraphics[width=1.\linewidth]{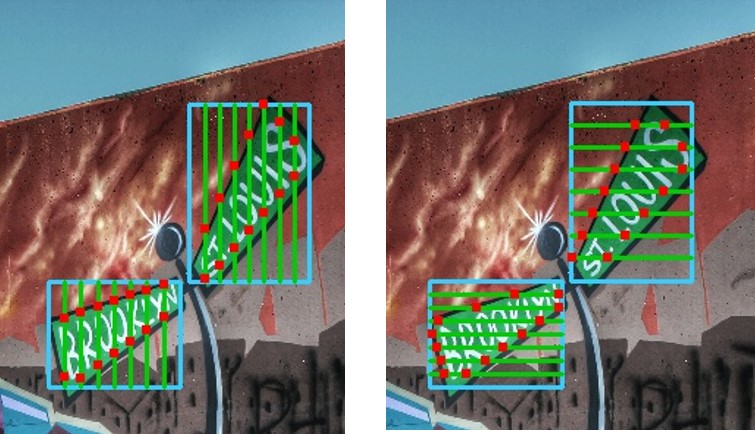}
    \end{center}
    \caption{Illustration of ground truth points generated by horizontally and vertically sliding lines.}
	\label{overview_img}
\end{figure}

1. We explore regressing multiple points on the border of text line, try to handle arbitrary-oriented and curve text detection based on Faster R-CNN and R-FCN.

2. We introduce a sliding line method to determine the ground truth points for the regression, and we make full use of the relevance of these points to generate more regular polygon.

\section{Related Work}

\begin{figure}
	\begin{center}
       \includegraphics[width=0.9\linewidth]{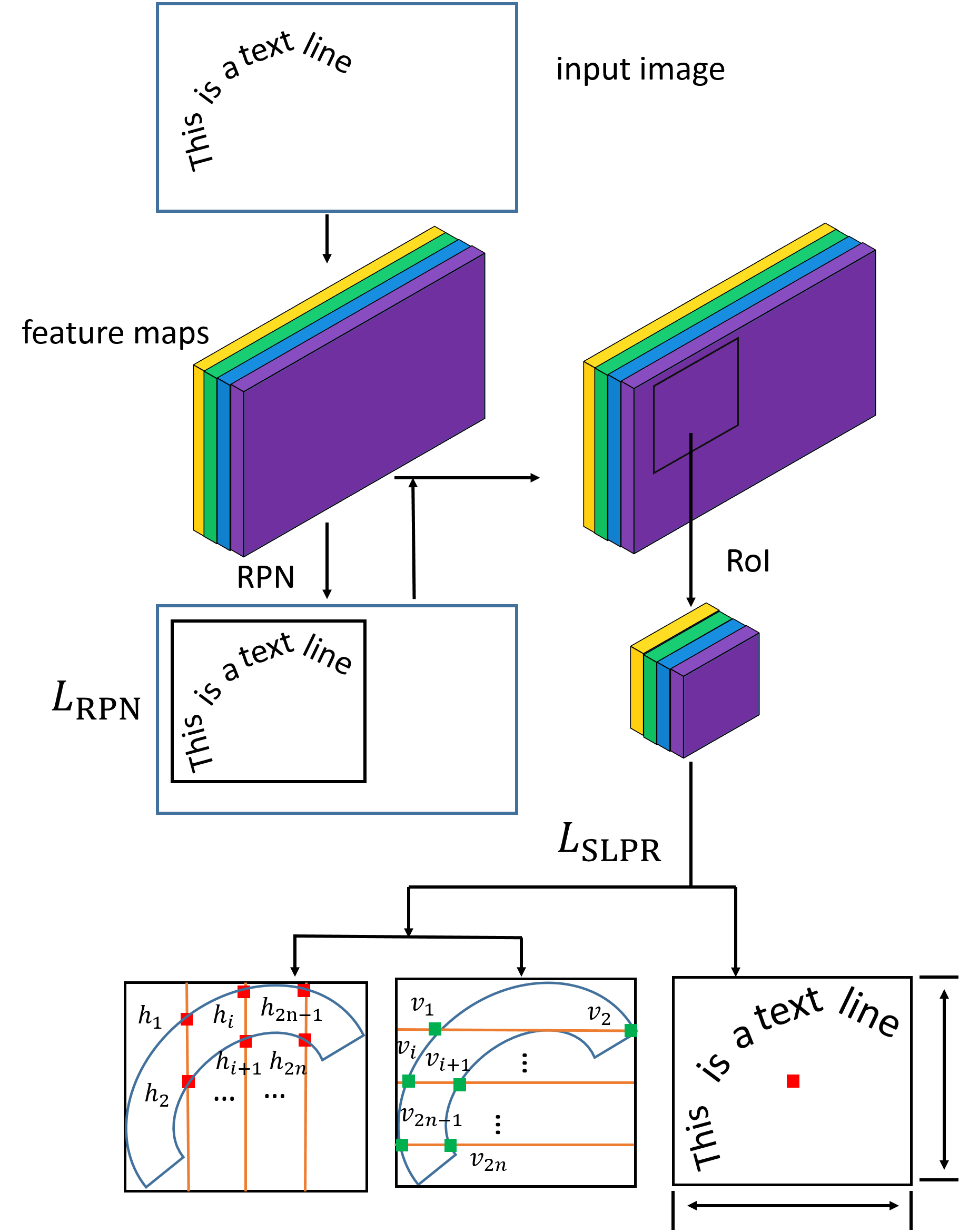}
    \end{center}
    \caption{The SLPR architecture.
    }
	\label{overview}
\end{figure}

In recent years, scene text detection and recognition has drawn more and more attention. But scene text detection remains a difficult problem due to its complicated orientation and background. All the methods can be divided into three categories: character based methods, word based methods and segmentation based methods. Character based methods often need synthetic datasets because labeling characters in text lines requires additional efforts. However, the generated data is greatly deviated from the real data, which can not make the trained model to achieve the state-of-the-art results on real dataset such as the popular ICDAR2015 Incidental Scene Text benchmark. In order to solve this problem, \cite{hu2017wordsup} used semi-supervised method to finetune model on real data and obtained good results.

The segmentation based methods have also been used in text detection recently. \cite{zhang2016multi} trained a fully convolutional network (FCN)~\cite{long2015fully,zhang2017watch} to predict the salient map of text regions, then traced the text line by combining the salient map and character components. \cite{wu2017self} added the border class to separate text from their neighbor. \cite{zhou2017east} and \cite{he2017deep} generated text maps and regress the size and angle of the corresponding quadrilateral, or coordinates of four vertexes at the same time. Compared with the traditional segmentation methods, they made a huge breakthrough on ICDAR2015 Incidental Scene Text benchmark.

Many methods of object detection can be applied to text detection, e.g., Faster R-CNN \cite{ren2017faster}, SSD \cite{liu2016ssd}, R-FCN \cite{dai2016r} and YOLO \cite{redmon2016you}. \cite{liao2017textboxes} used irregular $1\times5$ convolutional filters instead of the standard $3\times3$ convolutional filters to make the network more suitable for long text detection. \cite{he2017single} used the attention map to remove background noise. Recently more and more researchers proposed 2-step methods based on Faster R-CNN or R-FCN. \cite{jiang2017r2cnn} firstly generated axis-aligned bounding boxes and then regressed the text quadrangle. They used multi-scale pool operations on the roipool layer. \cite{dai2017fused} tried to segment and detect text simultaneously. Considering the particularity of text line, \cite{ma2017arbitrary} appended different angle anchors which are suitable for arbitrary-oriented text line. More recently, \cite{kheng2017total} considered the polygon case and labeled a new dataset of curve text. \cite{yuliang2017detecting} also constructed a curve text dataset named CTW1500, and they proposed a new structure named curve text detector (CTD) to solve curve text detection problem.

\section{Method}

Our model can be applied to any 2-step object detection framework such as Faster R-CNN and R-FCN. Our system simultaneously regresses the minimum rectangle including text line and the coordinates of some specific points on the boundary of text line. More specifically, take Faster R-CNN as an example, we first get some interesting regions using the RPN, then we not only regress the position of the rectangle, but also regress the coordinates of the points on the edge of the text line, finally we can get arbitrary shape text area.

\begin{figure}
  \begin{center}
       \includegraphics[width=1.\linewidth]{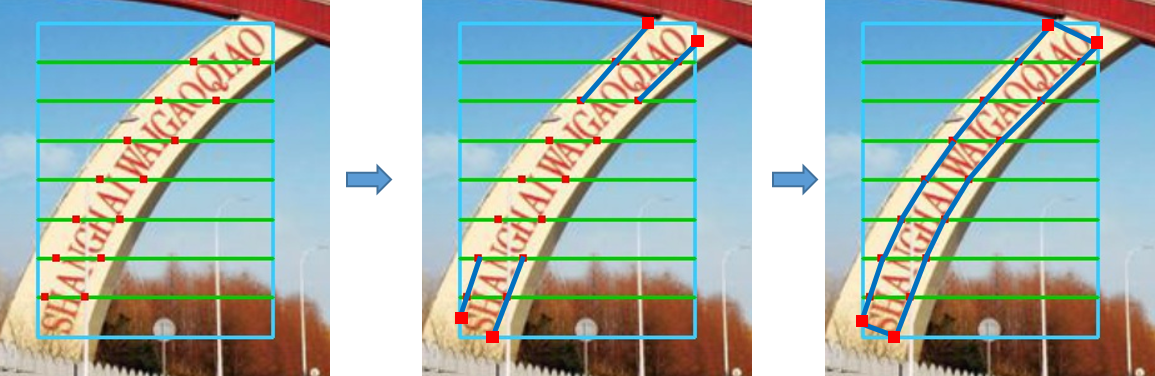}
    \end{center}
    \caption{Restoration of the polygon by using intersection points on sliding lines along the long side.
     }
  \label{restore_h}
\end{figure}

\begin{figure}
  \begin{center}
       \includegraphics[width=1.\linewidth]{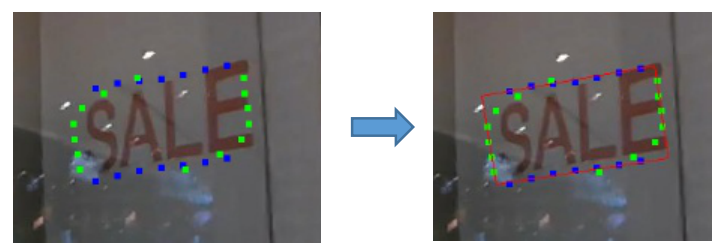}
    \end{center}
    \caption{Restoration of the polygon by using all intersection points from SLPR.}
  \label{restore_1}
\end{figure}

\subsection{Which points should be regressed? }

Obviously, how to determine the point set for restoring the polygon is quite important. We believe the simpler the rules, the easier the neural net learns. We do not regress the fixed points such as vertexes on the polygon because there are a large variety of shapes and angles in natural scene and it is difficult to define the order of fixed feature points for all shapes. Although for quadrilateral, we can perfectly restore it by regressing the corresponding four vertices, the determination of the order of four vertices requires a complicated rule which is difficult for the neural net to learn. Alternatively, as shown in Fig. \ref{overview}, we introduce some rules to vertically and horizontally slide the lines (we use equidistant sliding in our experiment) on text line and then regress the intersection of sliding lines and text line border. On the other hand, the correlation exists among the coordinates of different intersection points due to the constraints of the sliding lines. It is not necessary to regress both x-coordinate and y-coordinate of all points simultaneously. If it is horizontal sliding, the x-coordinate of the point on the text boundary can be calculated by the coordinates of the rectangle, so we only need to regress the y-coordinate of these points. Similarly, if it is vertical sliding, we only need to regress the x-coordinate of these points. This method not only reduces the computational complexity of the network, but also adds restraints to the regressed points as the prior knowledge which can prevent generating polygons with weird shapes and further improve the accuracy. As for the number of sliding lines, we observe that this parameter is not sensitive to quadrangle text line. But in order to restore other shape text line well, after balancing the performance and network complexity, seven sliding lines are used for we decided to for vertical and horizontal directions, respectively. Accordingly a total of 14 lines with 28 intersection points are generated.

\subsection{The multi-task learning}
To optimize the neural network parameters, as illustrated in Fig. \ref{overview}, we adopt the multi-task learning to define the loss function $L$ as:
\begin{align}
L=L_{\text{RPN}}+L_{\text{SLPR}}
\end{align}
\begin{align}L_{\text{RPN}}=L_{\text{RCLS}}+\lambda_{\text{R}}L_{\text{RB}}\end{align}
\begin{align}L_{\text{SLPR}}=L_{\text{CLS}}+\lambda_{\text{B}}L_{\text{B}}+\lambda_{\text{S}}L_{\text{SLPRB}}\end{align}
where $L_{\text{RPN}}$ is the region proposal loss, $L_{\text{RCLS}}$ is region proposal classification loss, $L_{\text{RB}}$ is box regression loss. $L_{\text{SLPR}}$ is the loss for the second step after RPN. Similarly, the first two items $L_{\text{CLS}}$ and $L_{\text{B}}$ are respectively classification loss and box regression loss. $\lambda_{\text{R}}$, $\lambda_{\text{B}}$ and $\lambda_{\text{S}}$ are the related weighting factors, which are all set to 1 in this study. $L_{\text{SLPRB}}$ is the proposed new loss item for SLPR:
\begin{align}\label{slprbox}
L_{\text{SLPRB}}= \frac{1}{4n} \left[ \sum_{j=1}^{2n}L_{\text{Reg}}(x_{v_{j}},x^{*}_{v_{j}})+\sum_{i=1}^{2n}L_{\text{Reg}}(y_{h_{i}},y^{*}_{h_{i}})\right] \end{align}
$L_{\text{Reg}}$ is the smooth L1 loss for the box regression task:

\begin{align}L_{\text{Reg}}(z,z^{*})=
\begin{cases}
0.5(z-z^{*})^{2}& \text{if} \ |z-z^{*}| < 1\\
|z-z^{*}|-0.5& \text{otherwise}
\end{cases}\end{align}

\begin{table*}
  \begin{center}

      \caption{The performance comparison of SLPR method  with different settings on ICDAR2015 Incidental Scene Text dataset.}
    \begin{tabular}{|c|c|c|c|c|c|}

      \hline
         Scales & NMS & Restoration& Precision (\%)& Recall (\%)& Hmean (\%)\\

      \hline\hline

      (850)      & PNMS&BHVP   &86.1 &81.6  &83.8\\
      (850)      & NMS&BHVP    &86.8 &80.1  &83.3\\
      (850)      & PNMS&PLS    &85.6 &81.5  &83.5\\
      (850,1000)  & PNMS& BHVP & 85.5 & 83.6 & 84.5 \\
      (850,1000)  & NMS& BHVP & 86.2 & 82.7 & 84.4 \\
      (850,1000)  & PNMS& PLS & 84.9 & 83.6 & 84.3 \\

      \hline

    \end{tabular}
  \label{table_mine}
  \end{center}

\end{table*}

In Eq.~(\ref{slprbox}), $n$ represents the number of sliding lines in one direction and we set $n=7$ in our experiments. In general, each line has two intersection points with the text line border. If there are more than two intersection points, we take the smallest and the largest coordinates. $x_{v_{j}}$ is x-coordinate of the intersection point $v_{j}$ of vertically sliding lines and text line border while $y_{h_{i}}$ is y-coordinate of the intersection point $h_{i}$ of horizontally sliding lines and text line border. $x^{*}_{v_{j}}$ and $y^{*}_{h_{i}}$ are the corresponding estimated points from neural net outputs. For horizontally sliding lines, we only regress the y-coordinate of its intersection point. For vertically sliding lines, we only regress the x-coordinate of its intersection point. The other coordinates can be restored through the coordinates of the rectangle:
\begin{align}y_{v_{j}}=y_{\text{min}}+(y_{\text{max}}-y_{\text{min}})\frac{\lfloor (j-1)/2 \rfloor+1}{(n+1)}\end{align}
\begin{align}x_{h_{i}}=x_{\text{min}}+(x_{\text{max}}-x_{\text{min}})\frac{\lfloor (i-1)/2 \rfloor+1}{(n+1)}\end{align}
$x_{\text{min}}$ and $y_{\text{min}}$ represent the minimum x-coordinate and y-coordinate of the rectangular border while $x_{\text{max}}$ and $y_{\text{max}}$ represent the maximum x-coordinate and y-coordinate of the rectangular border. $\lfloor \cdot \rfloor$ is the floor function. In a word, in order to regress the coordinates of polygon, 32 parameters should be considered including 4 parameters for the rectangle and 28 parameters to represent x and y coordinates of intersection points on text line border.

\subsection{Restoration of polygon}

Through the above SLPR method, we can obtain multiple points from the output of neural nets. To restore the final quadrilateral or polygon, the following two approaches are adopted and compared:

\subsubsection{Only Using Points in Long Side (PLS)}

The text line always extends to the long side, and the lines that slide along the long side can better reflect the shape of the text. In fact we can restore the polygon by only scanning the long side, as shown in Fig.~\ref{restore_h}. Specifically, we firstly judge whether the text line is horizontal or vertical through the regressed rectangle, and then restore the polygon through points in the corresponding direction. Taking the vertical direction as an example in Fig. \ref{restore_h}, since we do not regress the intersection point on the rectangular border, we firstly extend the four lines near the border to find four intersection points with the rectangle, then we connect four new points and other intersection points to generate polygon.

\subsubsection{Using Both of Horizontal and Vertical Points (BHVP)}

In fact, if we use both horizontal and vertical points to restore polygon, we can calculate a polygon or quadrangle that passes through these points roughly as shown in Fig.\ref{restore_1} by using the method in \cite{tensmeyer2017pagenet}. In this way we can obtain dense enough points in both horizontal and vertical direction and we do not need to calculate the intersection with the rectangle as in PLS method. However, we observe BHVP is not as effective as PLS for the polygon case. So we use this method only on the quadrilateral dataset (ICDAR2015 Incidental Scene Text).

\subsection{Polygonal non-maximum suppression}

Non-maximum suppression (NMS) is a basic method commonly used in the object detection, and its purpose is to remove duplicate boxes. The traditional NMS method is based on rectangular boxes, which is not the best choice for other shapes. In recent years, other NMS approaches were investigated, e.g., locality-aware NMS \cite{zhou2017east}, inclined NMS \cite{jiang2017r2cnn}, Mask-NMS \cite{dai2017fused} and polygonal NMS (PNMS) \cite{yuliang2017detecting}. As we consider the polygon in this study, both NMS and PNMS are compared in our experiment.

\begin{table}
	
	\begin{center}
			\caption{The comparison with the-state-of-the-art on ICDAR2015 Incidental Scene Text dataset.}
		\begin{tabular}{|l|c|c|c|}
			\hline
			   Methods  & Precision (\%)& Recall (\%)& Hmean (\%)\\

			\hline\hline
      HUST \cite{karatzas2015icdar} &44.0& 37.8 & 40.7 \\
      Zhang et al. \cite{zhang2016multi} &70.8&43.1  &53.6\\
      RRPN \cite{ma2017arbitrary} & 82.2& 73.2  & 77.4\\
      WordSup \cite{hu2017wordsup} &79.3&77.0  &78.2\\

			EAST \cite{zhou2017east} & 83.3& 78.3  &80.7 \\
			Deep direct regression \cite{he2017deep} & 82.0 & 80.0 &81.0 \\
      R$^{2}$CNN \cite{jiang2017r2cnn} & 85.6&  79.7 & 82.5\\
      FSTN \cite{dai2017fused} & {\bfseries 88.6} & 80.0 &84.1 \\

			\hline\hline

			SLPR (Ours) & 85.5 & {\bfseries83.6} & {\bfseries 84.5 }\\

			\hline
		
		\end{tabular}
	\label{table0}
	\end{center}
	
\end{table}

\section{Experiments}

\begin{figure*}%
    \centering
    \subfloat{{\includegraphics[scale=0.15]{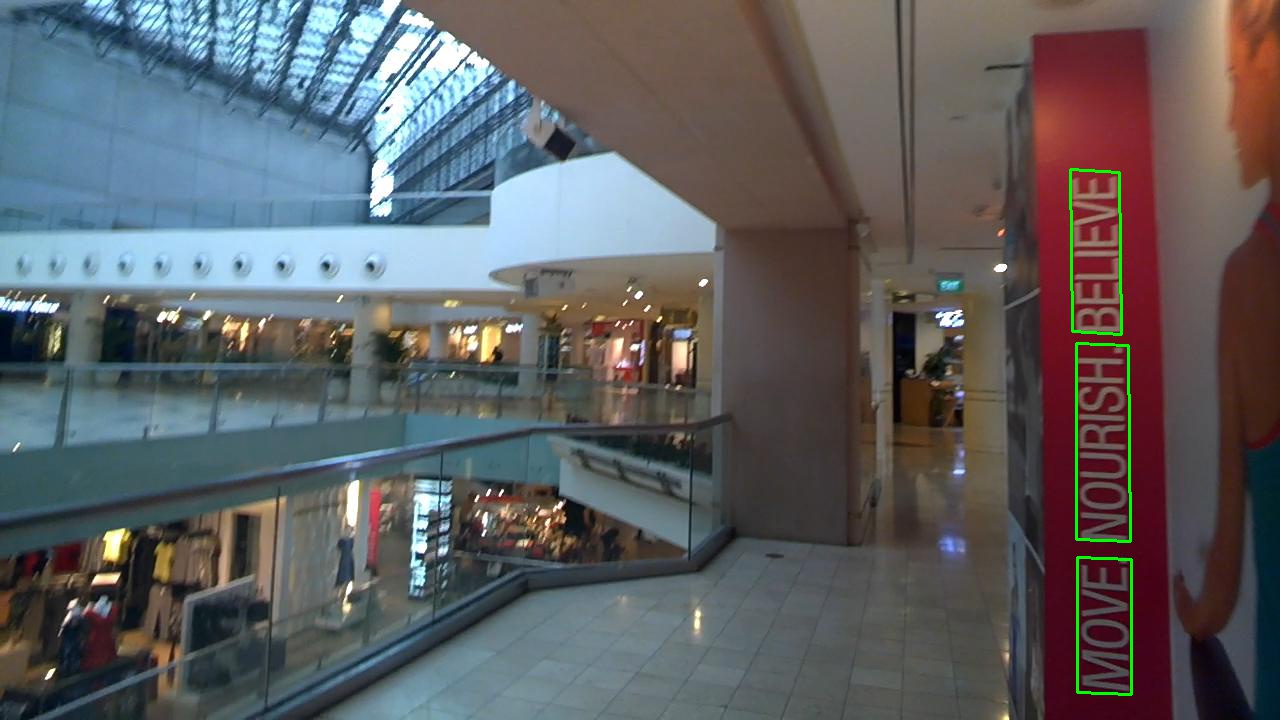} }}%
    \subfloat{{\includegraphics[scale=0.15]{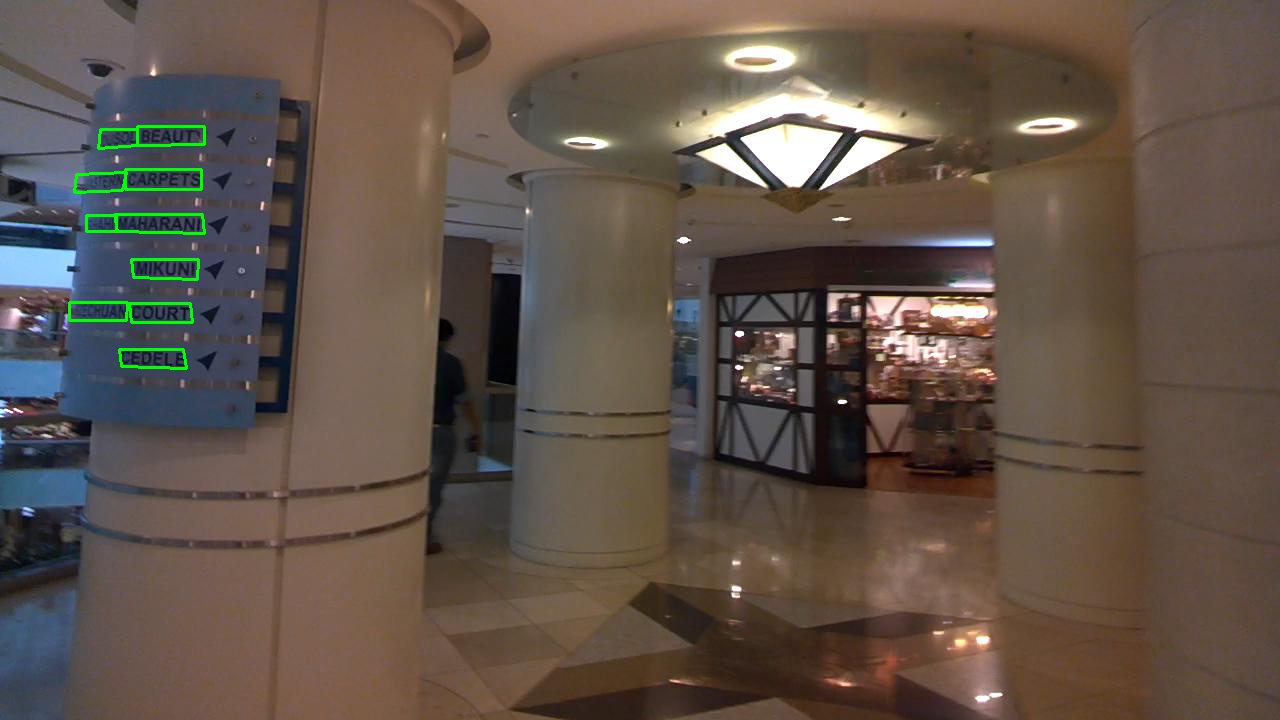} }}\\%
    \subfloat{{\includegraphics[scale=0.15]{test_img_154.jpg} }}%
    \subfloat{{\includegraphics[scale=0.15]{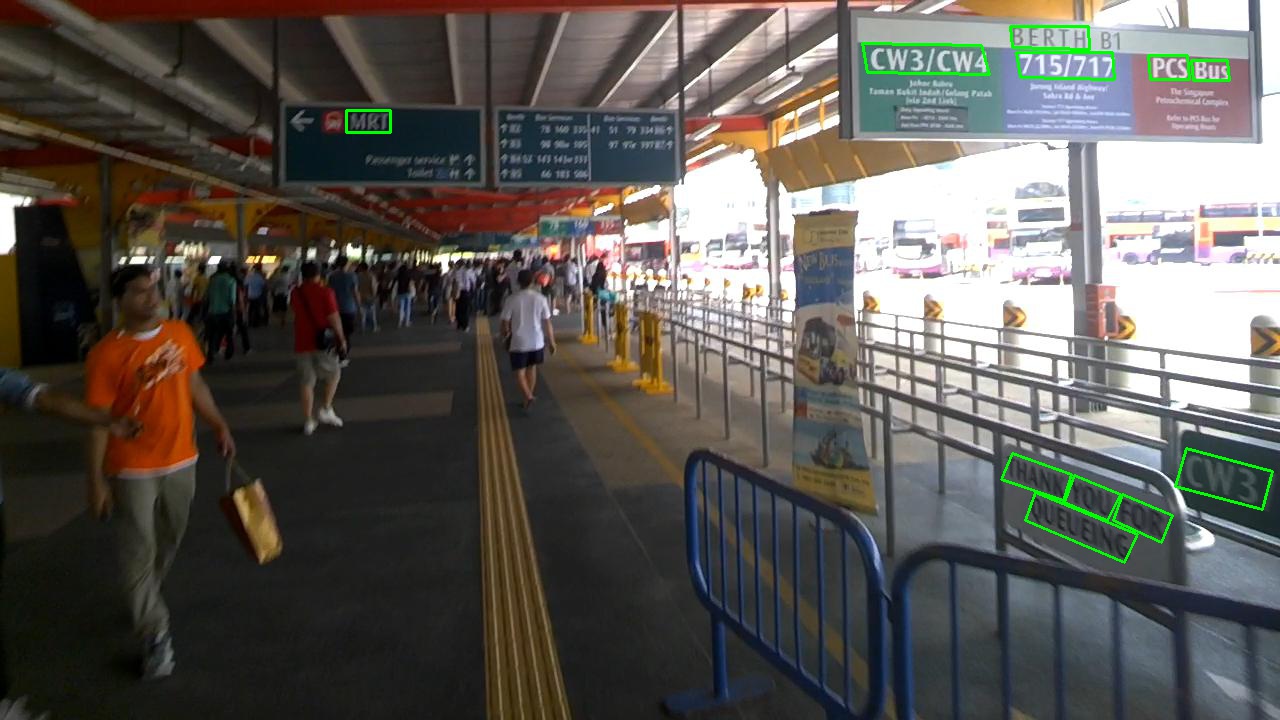} }}%
    \caption{The detection results on ICDAR2015 Incidental Scene Text dataset.}%
    \label{fig:icdar}%
\end{figure*}

\subsection{Datasets}
\subsubsection{ICDAR2015 Incidental Scene Text.}
ICDAR2015 Incidental Scene Text dataset \cite{karatzas2015icdar} is commonly used benchmark for detecting arbitrary-angle quadrangular text lines. It contains 1000 images for training, 500 images for testing. Some words which are too short, or unclear is annotated as don't cared samples.

\subsubsection{CTW1500}
Curve text dataset (CTW1500) is constructed by Yuliang et al. \cite{yuliang2017detecting}. Different from traditional text datasets, a text line is labelled by a polygon with 14 points.

\subsection{Implementation Details}

Since our proposed SLPR can be applied to any 2-step object detection framework. We adopted Faster R-CNN in ICDAR2015 Incidental Scene Text. And because \cite{yuliang2017detecting} also proposed a 2-step framework based on R-FCN while presenting the CTW1500 dataset, to perform a fair comparison, we directly used the network in \cite{yuliang2017detecting} from \url{https://github.com/Yuliang-Liu/Curve-Text-Detector}. All experiments were implemented in Caffe \cite{jia2014caffe} by using the NVIDIA GTX 1080Ti GPU.

\subsubsection{ICDAR2015 Incidental Scene Text.}
For Faster R-CNN structure, we used an additional $ 64 ^ {2} $ anchor and replaced RoIPool with RoIAlign \cite{he2017mask} because the text line is smaller than other objects. We set anchor scales as [$64^{2}, 128^{2}, 256^{2}, 512^{2}$] and set ratios as [0.5, 1, 2]. The base network is VGG16 \cite{simonyan2014very}, which is initialized by the pre-trained model on ImageNet database. We used stochastic gradient descent (SGD) with back-propagation and the maximum iteration was $20 \times 10^{4}$. Learning rates started from $10^{-3}$, decays to one-tenth every $5 \times 10^{4}$ iterations. We set weight decay as $0.0005$, and momentum as $0.9$. We used 1000 training incidental images in ICDAR2015 Incidental Scene Text \cite{karatzas2015icdar} and the 229 training images from ICDAR 2013 to train our network. In order to prevent over-fitting we employed data augmentation. Specifically, we randomly resized the images to $[720, 850, 960, 1200, 1400]$ where the numbers represent the length of the short side, and randomly rotated the images among $[0^\circ, 15^\circ, 30^\circ, ..., 360^\circ]$.

\subsubsection{CTW1500}
We used the curve text detector (CTD) network which is based R-FCN from \cite{yuliang2017detecting}. \cite{yuliang2017detecting} also added LSTM units named transverse and longitudinal offset connection (TLOC) to learn the correlation of points. But we removed it. As we only used PLS to restore polygon, Eq.~(\ref{slprbox}) was modified as:
\[
\begin{split}
L_{\text{SLPRB}}= \frac{1}{4n} \left[ \lambda_{hw}I(\dfrac{h}{w}>k)\sum_{j=1}^{2n}L_{\text{Reg}}(x_{v_{j}},x^{*}_{v_{j}})+ \right.\\
\left. I(\dfrac{h}{w} < \dfrac{1}{k})\sum_{i=1}^{2n}L_{\text{Reg}}(y_{h_{i}},y^{*}_{h_{i}}) \right]
\end{split}
\]
$I(z)$ equals to $1$ when $z$ is true, otherwise $0$. $h$ and $w$ are the height and width of the rectangle. Because most of the texts in this dataset are horizontal text line, to solve the imbalance between horizontal and vertical samples, we added $\lambda_{hw}=4$ to balance the losses between them. And when $h$ is close to $w$ , the text line may be judged as horizontal or vertical, so we set $k$ as $0.8$. The base network is ResNet-50 \cite{he2016deep}, which is initialized by the pre-trained model on ImageNet database. The max iteration was $8 \times 10^{4}$. The learning rate in this experiment was always $10^{-3}$. We set weight decay as $0.0005$, and momentum as $0.9$. To conduct a fair comparison, we only used the training set in CTW1500 to train our network and did not use data augmentation.

\subsection{Results}

\subsubsection{ICDAR2015 Incidental Scene Text}

Table~\ref{table_mine} shows the results of SLPR system with different settings. First, for the restoration of the quadrangle for the text region, BHVP using all the points can achieve better results than PLS using only the long-side points. Second, even we aim to detect the quadrangle in this dataset, PNMS still outperforms NMS. Finally, the use of multi-scale is one way to improve detection performance on different target sizes. We also test the multi-scale results of our system at (850, 1000), which yields about  1\% absolute improvement of Hmean measure. Fig.~\ref{fig:icdar} lists several challenging examples of detection results on ICDAR2015 Incidental Scene Text dataset. Table \ref{table0} gives the comparison of SLPR with state-of-the-art results on ICDAR2015 Incidental Scene Text. We can observe that our method achieved the competitive results on this dataset.

\subsubsection{CTW1500}

\begin{table}
	
	\begin{center}
		\caption{The performance comparison of SLPR method with different NMS settings on CTW1500 dataset.}
		\begin{tabular}{|l|c|c|c|}
			\hline
			NMS Method  & Precision (\%)& Recall (\%)& Hmean (\%)\\

			\hline\hline
			
			NMS0.1 & 81.2& 64.3 &71.8 \\
			NMS0.2 & 81.0& 68.7 &74.3 \\
			NMS0.3 & 80.1& 70.1 &{\bfseries 74.8} \\
			
			PNMS0.1 & 80.5& 69.5 &74.6 \\
			PNMS0.2 & 78.8& 70.4 &74.4 \\
			PNMS0.3& 75.6& 71.3 &73.4 \\

			\hline
			
		\end{tabular}
		\label{table_nms}
	\end{center}
	
\end{table}

\begin{table}
	
	\begin{center}
			\caption{The comparison with CTD on CTW1500 dataset.}
		\begin{tabular}{|l|c|c|c|}
			\hline
			   Method  & Precision (\%)& Recall (\%)& Hmean (\%)\\

			\hline\hline

            CTD+TLOC \cite{yuliang2017detecting}  & 77.4& 69.8  & 73.4\\
            CTD \cite{yuliang2017detecting} & 74.3& 65.2  &69.5\\

			\hline\hline

            SLPR (ours)& 80.1& 70.1 &{\bfseries 74.8} \\

			\hline
		
		\end{tabular}
	\label{table_ctd}
	\end{center}
	
\end{table}

\begin{figure}
  \begin{center}
       \includegraphics[width=1.0\linewidth]{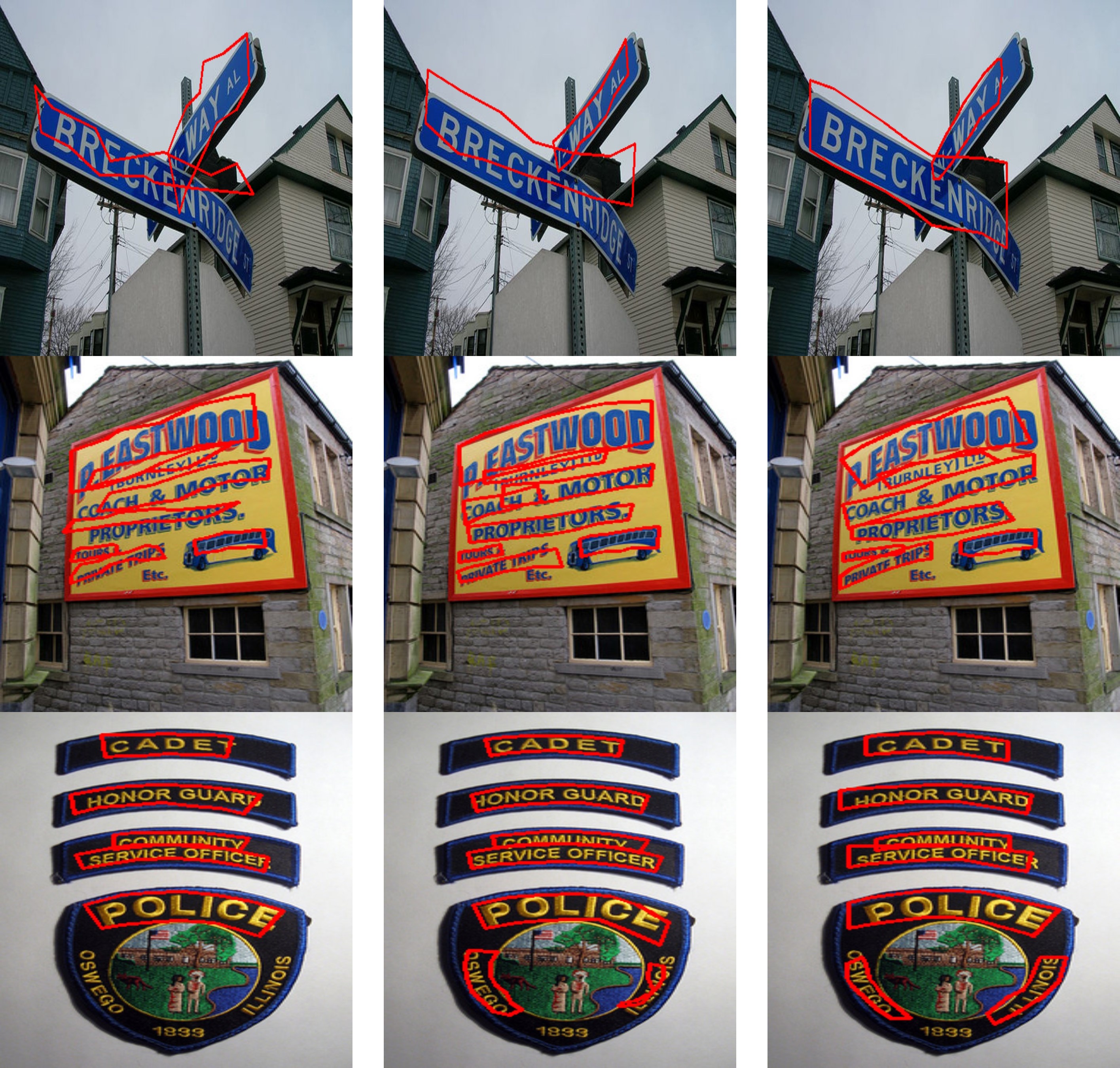}
    \end{center}
    \caption{The detection results on CTW1500 dataset. From left to right: CTD, CTD+TLOC and SLPR.
    }
  \label{result_ctd}
\end{figure}

Table~\ref{table_nms} shows the results of our method with different NMS settings. Different from the observation in ICDAR2015 Incidental Scene Text, our method achieved the best result on NMS0.3, namely the traditional NMS method with the threshold 0.3 for calculating the IoU (Intersection-over-Union). Table~\ref{table_ctd} lists the results of our method compared with CTD and CTD+TLOC. We removed TLOC from \cite{yuliang2017detecting} as our base network which is the same as CTD. Clearly, the Hmean performance of our SLPR method could be increased by $5.3\%$ over the CTD method, demonstrating the effectiveness of our simple rules to set the regression points. Even compared with the CTD+TLOC method with an additional LSTM network, SLPR still achieved $1.4\%$ improvement of Hmean performance. Fig.~\ref{result_ctd} gives several examples of the detection results of CTD, CTD+TLOC and our SLPR. We can observe that our method generated smoother regions and better detection results compared with CTD, which implied that the proposed SLPR can better handle the arbitrary-oriented case due to the novel design of the horizontally and vertically symmetrical scanning using sliding lines.

\section{Conclusion}

  In this study, we propose a novel SLPR method for the text detection in arbitrary-shape case. Compared with the curve text detection method CTD+TLOC \cite{yuliang2017detecting}, SLPR is more concise without using LSTM and obtains better performance. In the traditional quadrangle dataset (ICDAR2015 Incidental Scene Text), SLPR also achieves the state-of-the-art performance.

\section{Acknowledgment}
This work was supported in part by the National Key R\&D Program of China under contract No. 2017YFB1002202, in part by the National Natural Science Foundation of China under Grants 61671422 and U1613211, in part by the MOE-Microsoft Key Laboratory of USTC.






%

{
\bibliographystyle{IEEEtran}
\bibliography{refs}
}

\end{document}